# Generalized Singular Value Thresholding

**Canyi Lu[1], Changbo Zhu[1], Chunyan Xu[2], Shuicheng Yan[1], Zhouchen Lin[3,*]**
[1] Department of Electrical and Computer Engineering, National University of Singapore
[2] School of Computer Science and Technology, Huazhong University of Science and Technology
[3] Key Laboratory of Machine Perception (MOE), School of EECS, Peking University
canyilu@nus.edu.sg, zhuchangbo@gmail.com, xuchunyan01@gmail.com, eleyans@nus.edu.sg, zlin@pku.edu.cn

### Abstract
This work studies the Generalized Singular Value Thresholding (GSVT) operator $\mathbf{Prox}_g^\sigma(\cdot)$,

$$\mathbf{Prox}_g^\sigma(\mathbf{B}) = \arg\min_{\mathbf{X}} \sum_{i=1}^m g(\sigma_i(\mathbf{X})) + \frac{1}{2}||\mathbf{X} - \mathbf{B}||_F^2,$$

associated with a nonconvex function $g$ defined on the singular values of $\mathbf{X}$. We prove that GSVT can be obtained by performing the proximal operator of $g$ (denoted as $\mathbf{Prox}_g(\cdot)$) on the singular values since $\mathbf{Prox}_g(\cdot)$ is monotone when $g$ is lower bounded. If the nonconvex $g$ satisfies some conditions (many popular nonconvex surrogate functions, e.g., $\ell_p$-norm, $0 < p < 1$, of $\ell_0$-norm are special cases), a general solver to find $\mathbf{Prox}_g(b)$ is proposed for any $b \geq 0$. GSVT greatly generalizes the known Singular Value Thresholding (SVT) which is a basic subroutine in many convex low rank minimization methods. We are able to solve the nonconvex low rank minimization problem by using GSVT in place of SVT.

## Introduction

The *sparse* and *low rank* structures have received much attention in recent years. There have been many applications which exploit these two structures, such as face recognition (Wright et al. 2009), subspace clustering (Cheng et al. 2010; Liu et al. 2013b) and background modeling (Candès et al. 2011). To achieve sparsity, a principled approach is to use the convex $\ell_1$-norm. However, the $\ell_1$-minimization may be suboptimal, since the $\ell_1$-norm is a loose approximation of the $\ell_0$-norm and often leads to an over-penalized problem. This brings the attention back to the nonconvex surrogate by interpolating the $\ell_0$-norm and $\ell_1$-norm. Many nonconvex penalities have been proposed, including $\ell_p$-norm ($0 < p < 1$) (Frank and Friedman 1993), Smoothly Clipped Absolute Deviation (SCAD) (Fan and Li 2001), Logarithm (Friedman 2012), Minimax Concave Penalty (MCP) (Zhang and others 2010), Geman (Geman and Yang 1995) and Laplace (Trzasko and Manduca 2009). Their definitions are shown in Table 1. Numerical studies (Candès, Wakin, and Boyd 2008) have shown that the nonconvex optimization usually outperforms convex models.

*Corresponding author.

Table 1: Popular nonconvex surrogate functions of $\ell_0$-norm ($||\theta||_0$).

| Penalty | Formula $g(\theta), \theta \geq 0, \lambda > 0$ |
|---|---|
| $\ell_p$-norm | $\lambda\theta^p, 0 < p < 1$. |
| SCAD | $\begin{cases} \lambda\theta, & \text{if } \theta \leq \lambda, \\ \frac{-\theta^2 + 2\gamma\lambda\theta - \lambda^2}{2(\gamma-1)}, & \text{if } \lambda < \theta \leq \gamma\lambda, \\ \frac{\lambda^2(\gamma+1)}{2}, & \text{if } \theta > \gamma\lambda. \end{cases}$ |
| Logarithm | $\frac{\lambda}{\log(\gamma+1)}\log(\gamma\theta+1)$ |
| MCP | $\begin{cases} \lambda\theta - \frac{\theta^2}{2\gamma}, & \text{if } \theta < \gamma\lambda, \\ \frac{1}{2}\gamma\lambda^2, & \text{if } \theta \geq \gamma\lambda. \end{cases}$ |
| Geman | $\frac{\lambda\theta}{\theta+\gamma}$. |
| Laplace | $\lambda(1 - \exp(-\frac{\theta}{\gamma}))$. |

The low rank structure is an extension of sparsity defined on the singular values of a matrix. A principled way is to use the nuclear norm which is a convex surrogate of the rank function (Recht, Fazel, and Parrilo 2010). However, it suffers from the same suboptimal issue as the $\ell_1$-norm in many cases. Very recently, many popular nonconvex surrogate functions in Table 1 are extended on the singular values to better approximate the rank function (Lu et al. 2014). However, different from the convex optimization, the nonconvex low rank minimization is much more challenging than the nonconvex sparse minimization.

The Iteratively Reweighted Nuclear Norm (IRNN) method is proposed to solve the following nonconvex low rank minimization problem (Lu et al. 2014)

$$\min_{\mathbf{X}} F(\mathbf{X}) = \sum_{i=1}^m g(\sigma_i(\mathbf{X})) + h(\mathbf{X}), \quad (1)$$

where $\sigma_i(\mathbf{X})$ denotes the $i$-th singular value of $\mathbf{X} \in \mathbb{R}^{m \times n}$ (we assume $m \leq n$ in this work). $g : \mathbb{R}^+ \to \mathbb{R}^+$ is continuous, concave and nonincreasing on $[0, +\infty)$. Popular nonconvex surrogate functions in Table 1 are some examples. $h : \mathbb{R}^{m \times n} \to \mathbb{R}^+$ is the loss function which has Lipschitz continuous gradient. IRNN updates $\mathbf{X}^{k+1}$ by minimizing a surrogate function which upper bounds the objective function in (1). The surrogate function is constructed by linearizing $g$ and $h$ at $\mathbf{X}^k$, simultaneously. In theory, IRNN guarantees to decrease the objective function value of (1) in each iteration. However, it may decrease slowly since the upper

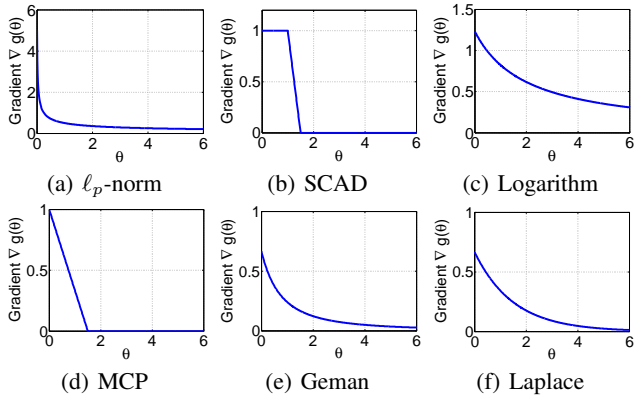

Figure 1: Gradients of some nonconvex functions (For $\ell_p$-norm, $p = 0.5$. For all penalties, $\lambda = 1$, $\gamma = 1.5$).

(a) $\ell_p$-norm  (b) SCAD  (c) Logarithm  (d) MCP  (e) Geman  (f) Laplace

bound surrogate may be quite loose. It is expected that minimizing a tighter surrogate will lead to a faster convergence.

A possible tighter surrogate function of the objective function in (1) is to keep $g$ and relax $h$ only. This leads to the following updating rule which is named as Generalized Proximal Gradient (GPG) method in this work

$$\mathbf{X}^{k+1} = \arg\min_{\mathbf{X}} \sum_{i=1}^{m} g(\sigma_i(\mathbf{X})) + h(\mathbf{X}^k)$$
$$+ \langle \nabla h(\mathbf{X}^k), \mathbf{X} - \mathbf{X}^k \rangle + \frac{\mu}{2}\|\mathbf{X} - \mathbf{X}^k\|_F^2$$
$$= \arg\min_{\mathbf{X}} \sum_{i=1}^{m} g(\sigma_i(\mathbf{X})) + \frac{\mu}{2}\|\mathbf{X} - \mathbf{X}^k + \frac{1}{\mu}\nabla h(\mathbf{X}^k)\|_F^2, \quad (2)$$

where $\mu > L(h)$, $L(h)$ is the Lipschitz constant of $h$, guarantees the convergence of GPG as shown later. It can be seen that solving (2) requires solving the following problem

$$\mathbf{Prox}_g^{\sigma}(\mathbf{B}) = \arg\min_{\mathbf{X}} \sum_{i=1}^{m} g(\sigma_i(\mathbf{X})) + \frac{1}{2}\|\mathbf{X} - \mathbf{B}\|_F^2. \quad (3)$$

In this work, the mapping $\mathbf{Prox}_g^{\sigma}(\cdot)$ is called the Generalized Singular Value Thresholding (GSVT) operator associated with the function $\sum_{i=1}^{m} g(\cdot)$ defined on the singular values. If $g(x) = \lambda x$, $\sum_{i=1}^{m} g(\sigma_i(\mathbf{X}))$ is degraded to the convex nuclear norm $\lambda\|\mathbf{X}\|_*$. Then (3) has a closed form solution $\mathbf{Prox}_g^{\sigma}(\mathbf{B}) = \mathbf{U}\mathrm{Diag}(\mathcal{D}_\lambda(\boldsymbol{\sigma}(\mathbf{B})))\mathbf{V}^T$, where $\mathcal{D}_\lambda(\boldsymbol{\sigma}(\mathbf{B})) = \{(\sigma_i(\mathbf{B}) - \lambda)_+\}_{i=1}^{m}$, and $\mathbf{U}$ and $\mathbf{V}$ are from the SVD of $\mathbf{B}$, i.e., $\mathbf{B} = \mathbf{U}\mathrm{Diag}(\boldsymbol{\sigma}(\mathbf{B}))\mathbf{V}^T$. This is the known Singular Value Thresholding (SVT) operator associated with the convex nuclear norm (when $g(x) = \lambda x$) (Cai, Candès, and Shen 2010). More generally, for a convex $g$, the solution to (3) is

$$\mathbf{Prox}_g^{\sigma}(\mathbf{B}) = \mathbf{U}\mathrm{Diag}(\mathbf{Prox}_g(\boldsymbol{\sigma}(\mathbf{B})))\mathbf{V}^T, \quad (4)$$

where $\mathbf{Prox}_g(\cdot)$ is defined element-wise as follows,

$$\mathbf{Prox}_g(b) = \arg\min_{x \geq 0} f_b(x) = g(x) + \frac{1}{2}(x-b)^2, ^1 \quad (5)$$

---
[1] For $x < 0$, $g(x) = g(-x)$. If $b \geq 0$, $\mathbf{Prox}_g(b) \geq 0$. If $b < 0$, $\mathbf{Prox}_g(b) = -\mathbf{Prox}_g(-b)$. So we only need to discuss the case $b, x \geq 0$ in this work.

where $\mathbf{Prox}_g(\cdot)$ is the known proximal operator associated with a convex $g$ (Combettes and Pesquet 2011). That is to say, solving (3) is equivalent to performing $\mathbf{Prox}_g(\cdot)$ on each singular value of $\mathbf{B}$. In this case, the mapping $\mathbf{Prox}_g(\cdot)$ is unique, i.e., (5) has a unique solution. More importantly, $\mathbf{Prox}_g(\cdot)$ is *monotone*, i.e., $\mathbf{Prox}_g(x_1) \geq \mathbf{Prox}_g(x_2)$ for any $x_1 \geq x_2$. This guarantees to preserve the nonincreasing order of the singular values after shrinkage and thresholding by the mapping $\mathbf{Prox}_g(\cdot)$. For a nonconvex $g$, we still call $\mathbf{Prox}_g(\cdot)$ as the proximal operator, but note that such a mapping may not be unique. It is still an open problem whether $\mathbf{Prox}_g(\cdot)$ is monotone or not for a nonconvex $g$. Without proving the monotonicity of $\mathbf{Prox}_g(\cdot)$, one cannot simply perform it on the singular values of $\mathbf{B}$ to obtain the solution to (3) as SVT. Even if $\mathbf{Prox}_g(\cdot)$ is monotone, since it is not unique, one also needs to carefully choose the solution $p_i \in \mathbf{Prox}_g(\sigma_i(\mathbf{B}))$ such that $p_1 \geq p_2 \geq \cdots \geq p_m$. Another challenging problem is that there does not exist a general solver to (5) for a general nonconvex $g$.

It is worth mentioning that some previous works studied the solution to (3) for some special choices of nonconvex $g$ (Nie, Huang, and Ding 2012; Chartrand 2012; Liu et al. 2013a). However, none of their proofs was rigorous since they ignored proving the monotone property of $\mathbf{Prox}_g(\cdot)$. See the detailed discussions in the next section. Another recent work (Gu et al. 2014) considered the following problem related to the weighted nuclear norm:

$$\min_{\mathbf{X}} f_{\mathbf{w},\mathbf{B}}(\mathbf{X}) = \sum_{i=1}^{m} w_i \sigma_i(\mathbf{X}) + \frac{1}{2}\|\mathbf{X} - \mathbf{B}\|_F^2, \quad (6)$$

where $w_i \geq 0$, $i = 1, \cdots, m$. Problem (6) is a little more general than (3) by taking different $g_i(x) = w_i x$. It is claimed in (Gu et al. 2014) that the solution to (6) is

$$\mathbf{X}^* = \mathbf{U}\mathrm{Diag}(\{\mathbf{Prox}_{g_i}(\sigma_i(\mathbf{B})), i = 1, \cdots, m\})\mathbf{V}^T, \quad (7)$$

where $\mathbf{B} = \mathbf{U}\mathrm{Diag}(\boldsymbol{\sigma}(\mathbf{B}))\mathbf{V}^T$ is the SVD of $\mathbf{B}$, and $\mathbf{Prox}_{g_i}(\sigma_i(\mathbf{B})) = \max\{\sigma_i(\mathbf{B}) - w_i, 0\}$. However, such a result and their proof are not correct. A counterexample is as follows:

$$\mathbf{B} = \begin{bmatrix} 0.0941 & 0.4201 \\ 0.5096 & 0.0089 \end{bmatrix}, \quad \mathbf{w} = \begin{bmatrix} 0.5 \\ 0.25 \end{bmatrix},$$

$$\mathbf{X}^* = \begin{bmatrix} -0.0345 & 0.1287 \\ 0.0542 & -0.0512 \end{bmatrix}, \quad \widehat{\mathbf{X}} = \begin{bmatrix} 0.0130 & 0.1938 \\ 0.1861 & -0.0218 \end{bmatrix},$$

where $\mathbf{X}^*$ is obtained by (7). The solution $\mathbf{X}^*$ is not optimal to (6) since there exists $\widehat{\mathbf{X}}$ shown above such that $f_{\mathbf{w},\mathbf{B}}(\widehat{\mathbf{X}}) = 0.2262 < f_{\mathbf{w},\mathbf{B}}(\mathbf{X}^*) = 0.2393$. The reason behind is that

$$(\mathbf{Prox}_{g_i}(\sigma_i(\mathbf{B})) - \mathbf{Prox}_{g_j}(\sigma_j(\mathbf{B})))(\sigma_i(\mathbf{B}) - \sigma_j(\mathbf{B})) \geq 0, \quad (8)$$

does not guarantee to hold for any $i \neq j$. Note that (8) holds when $0 \leq w_1 \leq \cdots \leq w_m$, and thus (7) is optimal to (6) in this case.

In this work, we give the first rigorous proof that $\mathbf{Prox}_g(\cdot)$ is monotone for any lower bounded function (regardless of the convexity of $g$). Then solving (3) is degenerated to solving (5) for each $b = \sigma_i(\mathbf{B})$. The Generalized Singular Value Thresholding (GSVT) operator $\mathbf{Prox}_g^{\sigma}(\cdot)$ associated with any lower bounded function in (3) is much more general than the known SVT associated with the convex nuclear norm (Cai, Candès, and Shen 2010). In order to compute GSVT, we analyze the solution to (5) for certain types of $g$ (some special cases are shown in Table 1) in theory, and propose a general solver to (5). At last, with GSVT, we can solve (1) by the Generalized Proximal Gradient (GPG) algorithm shown in (2). We test both Iteratively Reweighted Nuclear Norm (IRNN) and GPG on the matrix completion problem. Both synthesis and real data experiments show that GPG outperforms IRNN in terms of the recovery error and the objective function value.

## Generalized Singular Value Thresholding

### Problem Reformulation

A main goal of this work is to compute GSVT (3), and uses it to solve (1). We will show that, if $\mathbf{Prox}_g(\cdot)$ is monotone, problem (3) can be reformulated into an equivalent problem which is much easier to solve.

**Lemma 1.** *(von Neumann's trace inequality (Rhea 2011)) For any matrices $\mathbf{A}, \mathbf{B} \in \mathbb{R}^{m \times n}$ ($m \leq n$), $\mathrm{Tr}(\mathbf{A}^T\mathbf{B}) \leq \sum_{i=1}^m \sigma_i(\mathbf{A})\sigma_i(\mathbf{B})$, where $\sigma_1(\mathbf{A}) \geq \sigma_2(\mathbf{A}) \geq \cdots \geq 0$ and $\sigma_1(\mathbf{B}) \geq \sigma_2(\mathbf{B}) \geq \cdots \geq 0$ are the singular values of $\mathbf{A}$ and $\mathbf{B}$, respectively. The equality holds if and only if there exist unitaries $\mathbf{U}$ and $\mathbf{V}$ such that $\mathbf{A} = \mathbf{U}\mathrm{Diag}(\boldsymbol{\sigma}(\mathbf{A}))\mathbf{V}^T$ and $\mathbf{B} = \mathbf{U}\mathrm{Diag}(\boldsymbol{\sigma}(\mathbf{B}))\mathbf{V}^T$ are the SVDs of $\mathbf{A}$ and $\mathbf{B}$, simultaneously.*

**Theorem 1.** *Let $g: \mathbb{R}^+ \to \mathbb{R}^+$ be a function such that $\mathbf{Prox}_g(\cdot)$ is monotone. Let $\mathbf{B} = \mathbf{U}\mathrm{Diag}(\boldsymbol{\sigma}(\mathbf{B}))\mathbf{V}^T$ be the SVD of $\mathbf{B} \in \mathbb{R}^{m \times n}$. Then an optimal solution to (3) is*

$$\mathbf{X}^* = \mathbf{U}\mathrm{Diag}(\boldsymbol{\varrho}^*)\mathbf{V}^T, \quad (9)$$

*where $\boldsymbol{\varrho}^*$ satisfies $\varrho_1^* \geq \varrho_2^* \geq \cdots \geq \varrho_m^*$, $i = 1, \cdots, m$, and*

$$\varrho_i^* \in \mathbf{Prox}_g(\sigma_i(\mathbf{B})) = \operatorname*{argmin}_{\varrho_i \geq 0} g(\varrho_i) + \frac{1}{2}(\varrho_i - \sigma_i(\mathbf{B}))^2. \quad (10)$$

*Proof.* Denote $\sigma_1(\mathbf{X}) \geq \cdots \geq \sigma_m(\mathbf{X}) \geq 0$ as the singular values of $\mathbf{X}$. Problem (3) can be rewritten as

$$\min_{\boldsymbol{\varrho}: \varrho_1 \geq \cdots \geq \varrho_m \geq 0} \left\{ \min_{\boldsymbol{\sigma}(\mathbf{X}) = \boldsymbol{\varrho}} \sum_{i=1}^m g(\varrho_i) + \frac{1}{2}\|\mathbf{X} - \mathbf{B}\|_F^2 \right\}. \quad (11)$$

By using the von Neumann's trace inequality in Lemma 1, we have

$$\|\mathbf{X} - \mathbf{B}\|_F^2 = \mathrm{Tr}(\mathbf{X}^T\mathbf{X}) - 2\mathrm{Tr}(\mathbf{X}^T\mathbf{B}) + \mathrm{Tr}(\mathbf{B}^T\mathbf{B})$$

$$= \sum_{i=1}^m \sigma_i^2(\mathbf{X}) - 2\mathrm{Tr}(\mathbf{X}^T\mathbf{B}) + \sum_{i=1}^m \sigma_i^2(\mathbf{B})$$

$$\geq \sum_{i=1}^m \sigma_i^2(\mathbf{X}) - 2\sum_{i=1}^m \sigma_i(\mathbf{X})\sigma_i(\mathbf{B}) + \sum_{i=1}^m \sigma_i^2(\mathbf{B})$$

$$= \sum_{i=1}^m (\sigma_i(\mathbf{X}) - \sigma_i(\mathbf{B}))^2.$$

Note that the above equality holds when $\mathbf{X}$ admits the singular value decomposition $\mathbf{X} = \mathbf{U}\mathrm{Diag}(\boldsymbol{\sigma}(\mathbf{X}))\mathbf{V}^T$, where $\mathbf{U}$ and $\mathbf{V}$ are the left and right orthonormal matrices in the SVD of $\mathbf{B}$. In this case, problem (11) is reduced to

$$\min_{\boldsymbol{\varrho}: \varrho_1 \geq \cdots \geq \varrho_m \geq 0} \sum_{i=1}^m \left( g(\varrho_i) + \frac{1}{2}(\varrho_i - \sigma_i(\mathbf{B}))^2 \right). \quad (12)$$

Since $\mathbf{Prox}_g(\cdot)$ is monotone and $\sigma_1(\mathbf{B}) \geq \sigma_2(\mathbf{B}) \geq \cdots \geq \sigma_m(\mathbf{B})$, there exists $\varrho_i^* \in \mathbf{Prox}_g(\sigma_i(\mathbf{B}))$, such that $\varrho_1^* \geq \varrho_2^* \geq \cdots \geq \varrho_m^*$. Such a choice of $\varrho^*$ is optimal to (12), and thus (9) is optimal to (3). □

From the above proof, it can be seen that the monotone property of $\mathbf{Prox}_g(\cdot)$ is a key condition which makes problem (12) separable conditionally. Thus the solution (9) to (3) shares a similar formulation as the known Singular Value Thresholding (SVT) operator associated with the convex nuclear norm (Cai, Candès, and Shen 2010). Note that for a convex $g$, $\mathbf{Prox}_g(\cdot)$ is always monotone. Indeed,

$$(\mathbf{Prox}_g(b_1) - \mathbf{Prox}_g(b_2))(b_1 - b_2)$$
$$\geq (\mathbf{Prox}_g(b_1) - \mathbf{Prox}_g(b_2))^2 \geq 0, \forall b_1, b_2 \in \mathbb{R}^+.$$

The above inequality can be obtained by the optimality of $\mathbf{Prox}_g(\cdot)$ and the convexity of $g$.

The monotonicity of $\mathbf{Prox}_g(\cdot)$ for a nonconvex $g$ is still unknown. There were some previous works (Nie, Huang, and Ding 2012; Chartrand 2012; Liu et al. 2013a) claiming that the solution (9) is optimal to (3) for some special choices of nonconvex $g$. However, their results are not rigorous since the monotone property of $\mathbf{Prox}_g(\cdot)$ is not proved. Surprisingly, we find that the monotone property of $\mathbf{Prox}_g(\cdot)$ holds for any lower bounded function $g$.

**Theorem 2.** *For any lower bounded function $g$, its proximal operator $\mathbf{Prox}_g(\cdot)$ is monotone, i.e., for any $p_i^* \in \mathbf{Prox}_g(x_i)$, $i = 1, 2$, $p_1^* \geq p_2^*$, when $x_1 > x_2$.*

Note that it is possible that $\sigma_i(\mathbf{B}) = \sigma_j(\mathbf{B})$ for some $i < j$ in (10). Since $\mathbf{Prox}_g(\cdot)$ may not be unique, we need to choose $\varrho_i^* \in \mathbf{Prox}_g(\sigma_i(\mathbf{B}))$ and $\varrho_j^* \in \mathbf{Prox}_g(\sigma_j(\mathbf{B}))$ such that $\varrho_i^* \leq \varrho_j^*$. This is the only difference between GSVT and SVT.

### Proximal Operator of Nonconvex Function

So far, we have proved that solving (3) is equivalent to solving (5) for each $b = \sigma_i(\mathbf{B})$, $i = 1, \cdots, m$, for any lower bounded function $g$. For a nonconvex $g$, only for some special cases, the candidate solutions to (5) have a closed form (Gong et al. 2013). There does not exist a general solver for a more general nonconvex $g$. In this section, we analyze the solution to (5) for a broad choice of the nonconvex $g$. Then a general solver will be proposed in the next section.

**Assumption 1.** *$g: \mathbb{R}^+ \to \mathbb{R}^+$, $g(0) = 0$. $g$ is concave, non-decreasing and differentiable. The gradient $\nabla g$ is convex.*

In this work, we are interested in the nonconvex surrogate of $\ell_0$-norm. Except the differentiablity of $g$ and the convexity of $\nabla g$, all the other assumptions in **Assumption** 1 are necessary to construct a surrogate of $\ell_0$-norm. As shown later, these two additional assumptions make our analysis much easier. Note that the assumptions for the nonconvex

1807

**Algorithm 1:** A General Solver to (5) in which $g$ satisfying **Assumption** 1

**Input**: $b \geq 0$.
**Output**: Identify an optimal solution, 0 or
$$\hat{x}^b = \max\{x|\nabla f_b(x) = 0, 0 \leq x \leq b\}.$$
**if** $\nabla g(b) = 0$ **then**
  return $\hat{x}^b = b$;
**else**
  // find $\hat{x}^b$ by fixed point iteration.
  $x_0 = b$. // Initialization.
  **while** not converge **do**
    $x_{k+1} = b - \nabla g(x_k)$;
    **if** $x_{k+1} < 0$ **then**
      return $\hat{x}^b = 0$;
      break;
    **end**
  **end**
**end**
Compare $f_b(0)$ and $f_b(\hat{x}^b)$ to identify the optimal one.

function considered in **Assumption** 1 are quite general. It is easy to verify that many popular surrogates of $\ell_0$-norm in Table 1 satisfy **Assumption** 1, including $\ell_p$-norm, Logarithm, MCP, Geman and Laplace penalties. Only the SCAD penalty violates the convex $\nabla g$ assumption, as shown in Figure 1.

**Proposition 1.** *Given $g$ satisfying **Assumption 1**, the optimal solution to (5) lies in $[0, b]$.*

The above fact is obvious since both $g(x)$ and $\frac{1}{2}(x-b)^2$ are nondecreasing on $[b, +\infty)$. Such a result limits the solution space, and thus is very useful for our analysis. Our general solver to (5) is also based on Proposition 1.

Note that the solutions to (5) lie in 0 or the local points $\{x|\nabla f_b(x) = \nabla g(x) + x - b = 0\}$. Our analysis is mainly based on the number of intersection points of $D(x) = \nabla g(x)$ and the line $C_b(x) = b - x$. Let $\bar{b} = \sup\{b \mid C_b(x) \text{ and } D(x) \text{ have no intersection}\}$. We have the solution to (5) in different cases. Please refer to the supplementary material for the detailed proofs.

**Proposition 2.** *Given $g$ satisfying **Assumption 1** and $\nabla g(0) = +\infty$. Restricted on $[0, +\infty)$, when $b > \bar{b}$, $C_b(x)$ and $D(x)$ have two intersection points, denoted as $P_1^b = (x_1^b, y_1^b)$, $P_2^b = (x_2^b, y_2^b)$, and $x_1^b < x_2^b$. If there does not exist $b > \bar{b}$ such that $f_b(0) = f_b(x_2^b)$, then $\mathbf{Prox}_g(b) = 0$ for all $b \geq 0$. If there exists $b > \bar{b}$ such that $f_b(0) = f_b(x_2^b)$, let $b^* = \inf\{b \mid f_b(0) = f_b(x_2^b)\}$. Then we have*

$$\mathbf{Prox}_g(b) = \underset{x \geq 0}{\operatorname{argmin}} f_b(x) \begin{cases} = x_2^b, & \text{if } b > b^*, \\ \ni 0, & \text{if } b \leq b^*. \end{cases}$$

**Proposition 3.** *Given $g$ satisfying **Assumption 1** and $\nabla g(0) < +\infty$. Restricted on $[0, +\infty)$, if we have $C_{\nabla g(0)}(x) = \nabla g(0) - x \leq \nabla g(x)$ for all $x \in (0, \nabla g(0))$, then $C_b(x)$ and $D(x)$ have only one intersection point*

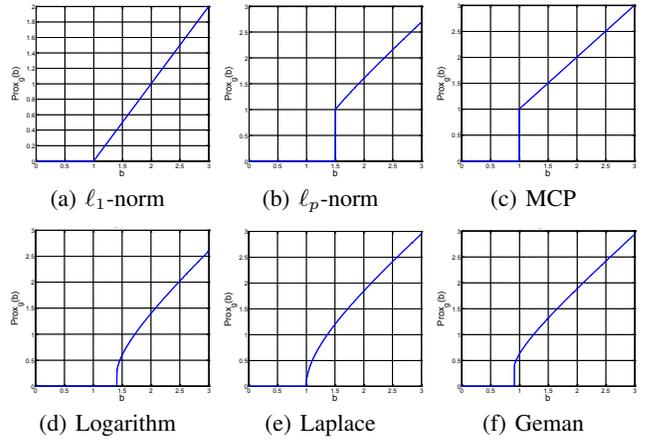

(a) $\ell_1$-norm  (b) $\ell_p$-norm  (c) MCP

(d) Logarithm  (e) Laplace  (f) Geman

Figure 2: Plots of $b$ v.s. $\mathbf{Prox}_g(b)$ for different choices of $g$: convex $\ell_1$-norm and popular nonconvex functions which satisfy **Assumption** 1 in Table 1.

*$(x^b, y^b)$ when $b > \nabla g(0)$. Furthermore,*

$$\mathbf{Prox}_g(b) = \underset{x \geq 0}{\operatorname{argmin}} f_b(x) \begin{cases} = x^b, & \text{if } b > \nabla g(0), \\ \ni 0, & \text{if } b \leq \nabla g(0). \end{cases}$$

*Suppose there exists $0 < \hat{x} < \nabla g(0)$ such that $C_{\nabla g(0)}(\hat{x}) = \nabla g(0) - \hat{x} > \nabla g(\hat{x})$. Then, when $\nabla g(0) \geq b > \bar{b}$, $C_b(x)$ and $D(x)$ have two intersection points, which are denoted as $P_1^b = (x_1^b, y_1^b)$ and $P_2^b = (x_2^b, y_2^b)$ such that $x_1^b < x_2^b$. When $\nabla g(0) < b$, $C_b(x)$ and $D(x)$ have only one intersection point $(x^b, y^b)$. Also, there exists $\tilde{b}$ such that $\nabla g(0) > \tilde{b} > \bar{b}$ and $f_{\tilde{b}}(0) = f_{\tilde{b}}(x_2^b)$. Let $b^* = \inf\{b \mid f_b(0) = f_b(x_2^b)\}$. We have*

$$\mathbf{Prox}_g(b) = \underset{x \geq 0}{\operatorname{argmin}} f_b(x) \begin{cases} = x^b, & \text{if } b > \nabla g(0), \\ = x_2^b, & \text{if } \nabla g(0) \geq b > b^*, \\ \ni 0, & \text{if } b \leq b^*. \end{cases}$$

**Corollary 1.** *Given $g$ satisfying **Assumption 1**. Denote $\hat{x}^b = \max\{x|\nabla f_b(x) = 0, 0 \leq x \leq b\}$ and $x^* = \arg\min_{x \in \{0, \hat{x}^b\}} f_b(x)$. Then $x^*$ is optimal to (5).*

The results in Proposition 2 and 3 give the solution to (5) in different cases, while Corollary 1 summarizes these results. It can be seen that one only needs to compute $\hat{x}^b$ which is the largest local minimum. Then comparing the objective function value at 0 and $\hat{x}^b$ leads to an optimal solution to (5).

## Algorithms

In this section, we first give a general solver to (5) in which $g$ satisfies **Assumption** 1. Then we are able to solve the GSVT problem (3). With GSVT, problem (1) can be solved by Generalized Proximal Gradient (GPG) algorithm as shown in (2). We also give the convergence guarantee of GPG.

### A General Solver to (5)

Given $g$ satisfying **Assumption** 1, as shown in Corollary 1, 0 and $\hat{x}^b = \max\{x|\nabla f_b(x) = 0, 0 \leq x \leq b\}$ are the candidate solutions to (5). The left task is to find $\hat{x}^b$ which is the largest local minimum point near $x = b$. So we can start

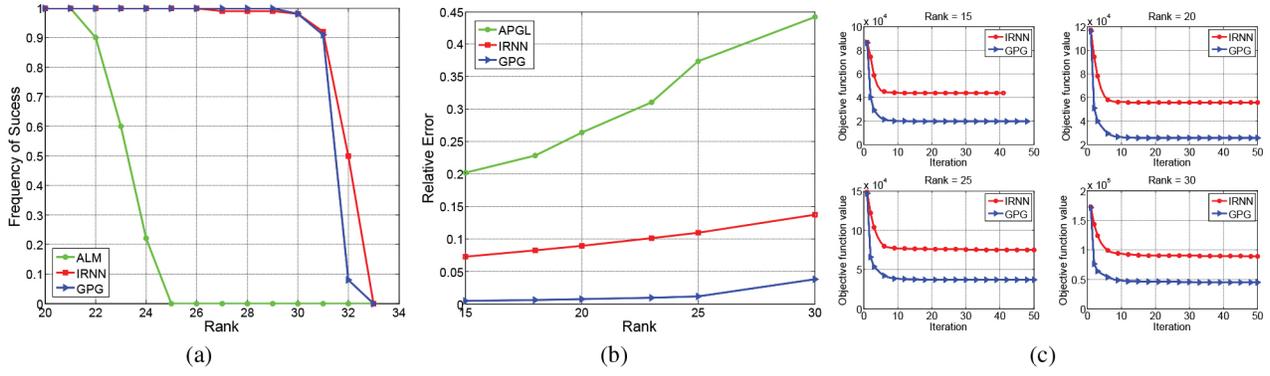

Figure 3: Experimental results of low rank matrix recovery on random data. (a) Frequency of Success (FoS) for a noise free case. (b) Relative error for a noisy case. (c) Convergence curves of IRNN and GPG for a noisy case.

searching for $\hat{x}^b$ from $x_0 = b$ by the fixed point iteration algorithm. Note that it will be very fast since we only need to search within $[0, b]$. The whole procedure to find $\hat{x}^b$ can be found in Algorithm 1. In theory, it can be proved that the fixed point iteration guarantees to find $\hat{x}^b$.

If $g$ is nonsmooth or $\nabla g$ is nonconvex, the fixed point iteration algorithm may also be applicable. The key is to find all the local solutions with smart initial points. Also all the nonsmooth points should be considered as the candidates.

All the nonconvex surrogates $g$ except SCAD in Table 1 satisfy **Assumption** 1, and thus the solution to (5) can be obtained by Algorithm 1. Figure 2 illustrates the shrinkage effect of proximal operators of these functions and the convex $\ell_1$-norm. The shrinkage and thresholding effect of these proximal operators are similar when $b$ is relatively small. However, when $b$ is relatively large, the proximal operators of the nonconvex functions are nearly unbiased, i.e., keeping $b$ nearly the same as the $\ell_0$-norm. On the contrast, the proximal operator of the convex $\ell_1$-norm is biased. In this case, the $\ell_1$-norm may be over-penalized, and thus may perform quite differently from the $\ell_0$-norm. This also supports the necessity of using nonconvex penalties on the singular values to approximate the rank function.

## Generalized Proximal Gradient Algorithm for (1)

Given $g$ satisfying **Assumption** 1, we are now able to get the optimal solution to (3) by (9) and Algorithm 1. Now we have a better solver than IRNN to solve (1) by the updating rule (2), or equivalently

$$\mathbf{X}^{k+1} = \mathbf{Prox}^{\sigma}_{\frac{1}{\mu}g}\left(\mathbf{X}^k - \frac{1}{\mu}\nabla h(\mathbf{X}^k)\right).$$

The above updating rule is named as Generalized Proximal Gradient (GPG) for the nonconvex problem (1), which generalizes some previous methods (Beck and Teboulle 2009; Gong et al. 2013). The main per-iteration cost of GPG is to compute an SVD, which is the same as many convex methods (Toh and Yun 2010a; Lin, Chen, and Ma 2009). In theory, we have the following convergence results for GPG.

**Theorem 3.** *If $\mu > L(h)$, the sequence $\{\mathbf{X}^k\}$ generated by (2) satisfies the following properties:*

*(1) $F(\mathbf{X}^k)$ is monotonically decreasing.*

*(2) $\lim_{k \to +\infty}(\mathbf{X}^k - \mathbf{X}^{k+1}) = \mathbf{0}$;*

*(3) If $F(\mathbf{X}) \to +\infty$ when $||\mathbf{X}||_F \to +\infty$, then any limit point of $\{\mathbf{X}^k\}$ is a stationary point.*

It is expected that GPG will decrease the objective function value faster than IRNN since it uses a tighter surrogate function. This will be verified by the experiments.

## Experiments

In this section, we conduct some experiments on the matrix completion problem to test our proposed GPG algorithm

$$\min_{\mathbf{X}} \sum_{i=1}^m g(\sigma_i(\mathbf{X})) + \frac{1}{2}||\mathcal{P}_\Omega(\mathbf{X}) - \mathcal{P}_\Omega(\mathbf{M})||_F^2, \quad (13)$$

where $\Omega$ is the index set, and $\mathcal{P}_\Omega : \mathbb{R}^{m \times n} \to \mathbb{R}^{m \times n}$ is a linear operator that keeps the entries in $\Omega$ unchanged and those outside $\Omega$ zeros. Given $\mathcal{P}_\Omega(\mathbf{M})$, the goal of matrix completion is to recover $\mathbf{M}$ which is of low rank. Note that we have many choices of $g$ which satisfies **Assumption** 1, and we simply test on the Logarithm penalty, since it is suggested in (Lu et al. 2014; Candès, Wakin, and Boyd 2008) that it usually performs well by comparing with other nonconvex penalties. Problem (13) can be solved by GPG by using GSVT (9) in each iteration. We compared GPG with IRNN on both synthetic and real data. The continuation technique is used to enhance the low rank matrix recovery in GPG. The initial value of $\lambda$ in the Logarithm penalty is set to $\lambda_0$, and dynamically decreased till reaching $\lambda_t$.

### Low-Rank Matrix Recovery on Random Data

We conduct two experiments on synthetic data without and with noises (Lu et al. 2014). For the noise free case, we generate $\mathbf{M} = \mathbf{M}_1\mathbf{M}_2$, where $\mathbf{M}_1 \in \mathbb{R}^{m \times r}$, $\mathbf{M}_2 \in \mathbb{R}^{r \times n}$ are i.i.d. random matrices, and $m = n = 150$. The underlying rank $r$ varies from 20 to 33. Half of the elements in $\mathbf{M}$ are missing. We set $\lambda_0 = 0.9||\mathcal{P}_\Omega(\mathbf{M})||_\infty$, and $\lambda_t = 10^{-5}\lambda_0$. The relative error RelErr$= ||\mathbf{X}^* - \mathbf{M}||_F/||\mathbf{M}||_F$ is used to evaluate the recovery performance. If RelErr is smaller than $10^{-3}$, $\mathbf{X}^*$ is regarded as a successful recovery of $\mathbf{M}$. We repeat the experiments 100 times for each $r$. We compare GPG by using GSVT with IRNN and the convex Augmented Lagrange Multiplier (ALM) (Lin, Chen, and Ma 2009). Figure 3 (a) plots $r$ v.s. the frequency of success. It can be seen

1809

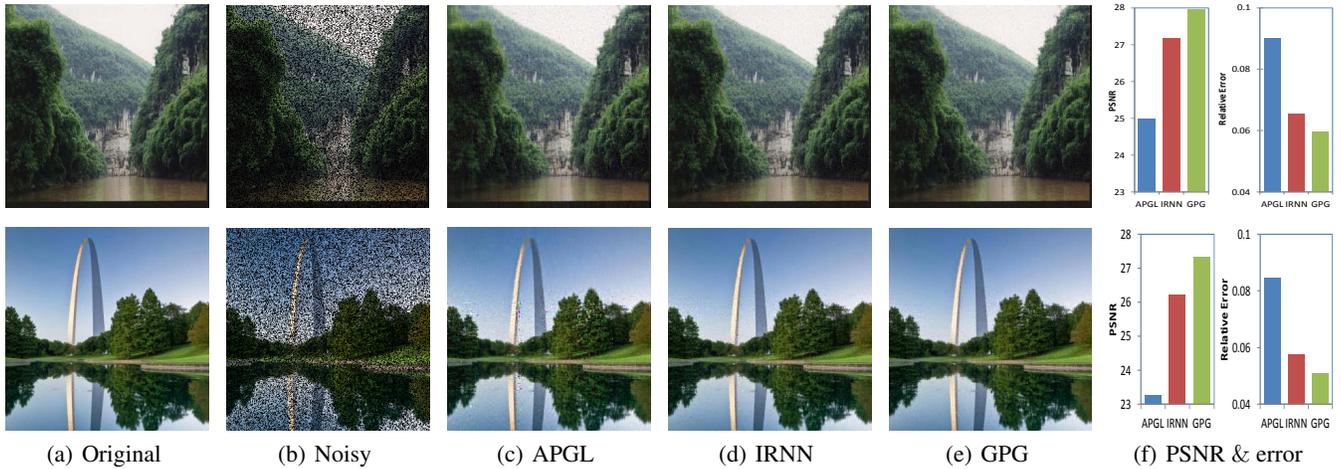

| (a) Original | (b) Noisy | (c) APGL | (d) IRNN | (e) GPG | (f) PSNR & error |

Figure 4: Image inpainting by APGL, IRNN, and GPG.

that GPG is slightly better than IRNN when $r$ is relatively small, while both IRNN and GPG fail when $r \geq 32$. Both of them outperform the convex ALM method, since the nonconvex logarithm penalty approximates the rank function better than the convex nuclear norm.

For the noisy case, the data matrix $\mathbf{M}$ is generated in the same way, but are added some additional noises $0.1\mathbf{E}$, where $\mathbf{E}$ is an i.i.d. random matrix. For this task, we set $\lambda_0 = 10||\mathcal{P}_\Omega(\mathbf{M})||_\infty$, and $\lambda_t = 0.1\lambda_0$ in GPG. The convex APGL algorithm (Toh and Yun 2010b) is compared in this task. Each method is run 100 times for each $r \in \{15, 18, 20, 23, 25, 30\}$. Figure 3 (b) shows the mean relative error. It can be seen that GPG by using GSVT in each iteration significantly outperforms IRNN and APGL. The reason is that $\lambda_t$ is not that small as in the noise free case. Thus, the upper bound surrogate of $g$ in IRNN will be much more loose than that in GPG. Figure 3 (c) plots some convergence curves of GPG and IRNN. It can be seen that GPG without relaxing $g$ will decrease the objective function value faster.

## Applications on Real Data

Matrix completion can be applied to image inpainting since the main information is dominated by the top singular values. For a color image, assume that 40% of pixels are uniformly missing. They can be recovered by applying low rank matrix completion on each channel (red, green and blue) of the image independently. Besides the relative error defined above, we also use the Peak Signal-to-Noise Ratio (PSNR) to evaluate the recovery performance. Figure 4 shows two images recovered by APGL, IRNN and GPG, respectively. It can be seen that GPG achieves the best performance, i.e., the largest PSNR value and the smallest relative error.

We also apply matrix completion for collaborative filtering. The task of collaborative filtering is to predict the unknown preference of a user on a set of unrated items, according to other similar users or similar items. We test on the MovieLens data set (Herlocker et al. 1999) which includes three problems, "movie-100K", "movie-1M" and "movie-10M". Since only the entries in $\Omega$ of $\mathbf{M}$ are known, we use Normalized Mean Absolute Error (NMAE) $||\mathcal{P}_\Omega(\mathbf{X}^*) - \mathcal{P}_\Omega(\mathbf{M})||_1/|\Omega|$ to evaluate the performance as in (Toh and Yun 2010b). As shown in Table 2, GPG achieves the best performance. The improvement benefits from the GPG algorithm which uses a fast and exact solver of GSVT (9).

Table 2: Comparison of NMAE of APGL, IRNN and GPG for collaborative filtering.

| Problem | size of $\mathbf{M}$: $(m, n)$ | APGL | IRNN | GPG |
|---|---|---|---|---|
| moive-100K | (943, 1682) | 2.76e-3 | 2.60e-3 | **2.53e-3** |
| moive-1M | (6040, 3706) | 2.66e-1 | 2.52e-1 | **2.47e-1** |
| moive-10M | (71567, 10677) | 3.13e-1 | 3.01e-1 | **2.89e-1** |

## Conclusions

This paper studied the Generalized Singular Value Thresholding (GSVT) operator associated with the nonconvex function $g$ on the singular values. We proved that the proximal operator of any lower bounded function $g$ (denoted as $\mathbf{Prox}_g(\cdot)$) is monotone. Thus, GSVT can be obtained by performing $\mathbf{Prox}_g(\cdot)$ on the singular values separately. Given $b \geq 0$, we also proposed a general solver to find $\mathbf{Prox}_g(b)$ for certain type of $g$. At last, we applied the generalized proximal gradient algorithm by using GSVT as the subroutine to solve the nonconvex low rank minimization problem (1). Experimental results showed that it outperformed previous method with smaller recovery error and objective function value.

For nonconvex low rank minimization, GSVT plays the same role as SVT in convex minimization. One may extend other convex low rank models to nonconvex cases, and solve them by using GSVT in place of SVT. An interesting future work is to solve the nonconvex low rank minimization problem with affine constraint by ALM (Lin, Chen, and Ma 2009) and prove the convergence.

## Acknowledgements

This research is supported by the Singapore National Research Foundation under its International Research Centre @Singapore Funding Initiative and administered by the IDM Programme Office. Z. Lin is supported by NSF China (grant nos. 61272341 and 61231002), 973 Program of China (grant no. 2015CB3525) and MSRA Collaborative Research Program. C. Lu is supported by the MSRA fellowship 2014.

# Supplementary Material of Generalized Singular Value Thresholding

**Canyi Lu**[1], **Changbo Zhu**[1], **Chunyan Xu**[2], **Shuicheng Yan**[1], **Zhouchen Lin**[3,*]

[1] Department of Electrical and Computer Engineering, National University of Singapore
[2] School of Computer Science and Technology, Huazhong University of Science and Technology
[3] Key Laboratory of Machine Perception (MOE), School of EECS, Peking University

canyilu@nus.edu.sg, zhuchangbo@gmail.com, xuchunyan01@gmail.com, eleyans@nus.edu.sg, zlin@pku.edu.cn

## 1 Ananlysis of the Proximal Operator of Nonconvex Function

In the following development, we consider the following problem

$$\mathbf{Prox}_g(b) = \arg\min_{x \geq 0} f_b(x) = g(x) + \frac{1}{2}(x-b)^2, \tag{1}$$

where $g(x)$ satisfies the following assumption.

**Assumption 1.** $g : \mathbb{R}^+ \to \mathbb{R}^+$, $g(0) = 0$. $g$ is concave, nondecreasing and differentiable. The gradient $\nabla g$ is convex.

Set $C_b(x) = b - x$ and $D(x) = \nabla g(x)$. Let $\bar{b} = \sup\{b \mid C_b(x)$ and $D(x)$ have no intersection$\}$, and $x_2^{\bar{b}} = \inf\{\ x \mid (x,y)$ is the intersection point of $C_{\bar{b}}(x)$ and $D(x)\}$.

### 1.1 Proof of Proposition 2

**Proposition 2.** *Given $g$ satisfying **Assumption 1** and $\nabla g(0) = +\infty$. Restricted on $[0, +\infty)$, when $b > \bar{b}$, $C_b(x)$ and $D(x)$ have two intersection points, denoted as $P_1^b = (x_1^b, y_1^b)$, $P_2^b = (x_2^b, y_2^b)$, and $x_1^b < x_2^b$. If there does not exist $b > \bar{b}$ such that $f_b(0) = f_b(x_2^b)$, then $\mathbf{Prox}_g(b) = 0$ for all $b \geq 0$. If there exists $b > \bar{b}$ such that $f_b(0) = f_b(x_2^b)$, let $b^* = \inf\{b \mid b > \bar{b}, f_b(0) = f_b(x_2^b)\ \}$. Then we have*

$$\mathbf{Prox}_g(b) = \operatorname*{argmin}_{x \geq 0} f_b(x) \begin{cases} = x_2^b, & \text{if } b > b^*, \\ \ni 0, & \text{if } b \leq b^*. \end{cases} \tag{2}$$

**Remark:** When $b^*$ exists and $b > b^*$, because $D(x) = \nabla g(x)$ is convex and decreasing, we can conclude that $C_b(x)$ and $D(x)$ have exactly two intersection points. When $b \leq b^*$, $C_b(x)$ and $D(x)$ may have multiple intersection points.

*Proof.* When $b > \bar{b}$, since $\nabla f_b(x) = D(x) - C_b(x)$, we can easily see that $f_b$ is increasing on $(0, x_1^b)$, decreasing on $(x_1^b, x_2^b)$ and increasing on $(x_2^b, b)$. So, 0 and $x_2^b$ are two local minimum points of $f_b(x)$ on $[0, b]$.

Case 1 : If there exists $b > \bar{b}$ such that $f_b(0) = f_b(x_2^b)$, denote $b^* = \inf\{b \mid b > \bar{b}, f_b(0) = f_b(x_2^b)\ \}$.

First, we consider $b > b^*$. Let $b = b^* + \varepsilon$ for some $\varepsilon > 0$. We have

$$\begin{aligned}
&f_b(x_2^{b^*}) - f_b(0) \\
=& \frac{1}{2}(x_2^{b^*} - b^* - \varepsilon)^2 + g(x^*) - \frac{1}{2}(b^* + \varepsilon)^2 \\
=& \frac{1}{2}(x_2^{b^*} - b^*)^2 - \frac{1}{2}(b^*)^2 - \varepsilon(x_2^{b^*} - b^*) - \varepsilon b^* \\
=& f_{b^*}(x_2^{b^*}) - f_{b^*}(0) - \varepsilon x_2^* \\
=& -\varepsilon x_2^* < 0.
\end{aligned}$$

Since $f_b$ is decreasing on $[x_2^{b^*}, x_2^b]$, we conclude that $f_b(0) > f_b(x_2^{b^*}) \geq f_b(x_2^b)$. So, when $b > b^*$, $x_2^b$ is the global minimum of $f_b(x)$ on $[0, b]$.

---
*Corresponding author.

Second, we consider $\bar{b} < b \leq b^*$. We show that $f_b(0) \leq f_b(x_2^b)$ by contradiction. Suppose that there exists $b$ such that $f_b(0) > f_b(x_2^b)$. Since $f_{\bar{b}}$ is strictly increasing on $(0, x_2^{\bar{b}})$, we have $f_{\bar{b}}(x_2^{\bar{b}}) > f_{\bar{b}}(0)$. Because we have

$$\begin{cases} f_{\bar{b}}(x_2^{\bar{b}}) > f_{\bar{b}}(0), \\ f_b(x_2^b) < f_b(0), \end{cases}$$

by a direct computation, we get

$$\begin{cases} g(x_2^{\bar{b}}) - x_2^{\bar{b}} \nabla g(x_2^{\bar{b}}) - \frac{1}{2}(x_2^{\bar{b}})^2 > 0, \\ g(x_2^b) - x_2^b \nabla g(x_2^b) - \frac{1}{2}(x_2^b)^2 < 0. \end{cases}$$

According to the intermediate value theorem, there exists $\tilde{x}$ such that $x_2^{\bar{b}} < \tilde{x} < x_2^b$ and $g(\tilde{x}) - \tilde{x}\nabla g(\tilde{x}) - \frac{1}{2}(\tilde{x})^2 = 0$. Let $\tilde{b} = \nabla g(\tilde{x}) + \tilde{x}$. Then, $(\tilde{x}, \tilde{b} - \tilde{x})$ is the intersection point of $C_{\tilde{b}}(x)$ and $D(x)$ such that $f_{\tilde{b}}(\tilde{x}) = f_{\tilde{b}}(0)$. Since $x_2^{\bar{b}} < \tilde{x} < x_2^b$ and $\nabla g$ is convex and nonincreasing, we conclude that $\bar{b} < \tilde{b} < b \leq b^*$, which contradicts the minimality of $b^*$.

Also, when $b \leq \bar{b}$, we have $\nabla f_b(x) = D(x) - C_b(x) \geq 0$, because $D(x)$ is above $C_b(x)$. So, the global minimum of $f_b(x)$ on $[0, b]$ is 0.

Case 2 : Suppose for all $b^* > \bar{b}$, $f_{b^*}(0) \neq f_{b^*}(x_2^{b^*})$. Since $f_{\bar{b}}$ is increasing on $(0, x_2^{\bar{b}})$, we have $f_{\bar{b}}(x_2^{\bar{b}}) > f_{\bar{b}}(0)$. We now show that for all $b > \bar{b}$, $f_b(x_2^b) \geq f_b(0)$. Suppose this is not true and there exists $b$ such that $b > \bar{b}$ and $f_b(x_2^b) < f_b(0)$. Because we have

$$\begin{cases} f_{\bar{b}}(x_2^{\bar{b}}) > f_{\bar{b}}(0), \\ f_b(x_2^b) < f_b(0), \end{cases}$$

by a direct computation, we get

$$\begin{cases} g(x_2^{\bar{b}}) - x_2^{\bar{b}} \nabla g(x_2^{\bar{b}}) - \frac{1}{2}(x_2^{\bar{b}})^2 > 0, \\ g(x_2^b) - x_2^b \nabla g(x_2^b) - \frac{1}{2}(x_2^b)^2 < 0. \end{cases}$$

So, according to the intermediate value theorem, there exists $\tilde{x}$ such that $g(\tilde{x}) - \tilde{x}\nabla g(\tilde{x}) - \frac{1}{2}(\tilde{x})^2 = 0$. Let $\tilde{b} = \nabla g(\tilde{x}) + \tilde{x}$. Then, $(\tilde{x}, \tilde{b} - \tilde{x})$ is the intersection point of $C_{\tilde{b}}(x)$ and $D(x)$ such that $f_{\tilde{b}}(\tilde{x}) = f_{\tilde{b}}(0)$. Since $x_2^{\bar{b}} < \tilde{x} < x_2^b$ and $\nabla g$ is convex and nonincreasing, we conclude that $\bar{b} < \tilde{b} < b$, which contradicts $f_{b^*}(0) \neq f_{b^*}(x_2^{b^*})$ for all $b^* > \bar{b}$. So, for all $b > \bar{b}$, 0 is the minimum of $f_b(x)$ on $[0, b]$. Similarly, when $b \leq \bar{b}$, we have $\nabla f_b(x) = D(x) - C_b(x) \geq 0$, because $D(x)$ is above $C_b(x)$. So, the global minimum of $f_b(x)$ on $[0, b]$ is 0. The proof is completed. □

## 1.2 Proof of Proposition 3

**Proposition 3.** *Given $g$ satisfying **Assumption 1** and $\nabla g(0) < +\infty$. Restricted on $[0, +\infty)$, if we have $C_{\nabla g(0)}(x) = \nabla g(0) - x \leq \nabla g(x)$ for all $x \in (0, \nabla g(0))$, then $C_b(x)$ and $D(x)$ have only one intersection point $(x^b, y^b)$ when $b > \nabla g(0)$. Furthermore,*

$$\mathbf{Prox}_g(b) = \operatorname*{argmin}_{x \geq 0} f_b(x) \begin{cases} = x^b, & \text{if } b > \nabla g(0), \\ \ni 0, & \text{if } b \leq \nabla g(0). \end{cases} \tag{3}$$

*Suppose there exists $0 < \hat{x} < \nabla g(0)$ such that $C_{\nabla g(0)}(\hat{x}) = \nabla g(0) - \hat{x} > \nabla g(\hat{x})$. Then, when $\nabla g(0) \geq b > \bar{b}$, $C_b(x)$ and $D(x)$ have two intersection points, which are denoted as $P_1^b = (x_1^b, y_1^b)$ and $P_2^b = (x_2^b, y_2^b)$ such that $x_1^b < x_2^b$. When $\nabla g(0) < b$, $C_b(x)$ and $D(x)$ have only one intersection point $(x^b, y^b)$. Also, there exists $\tilde{b}$ such that $\nabla g(0) > \tilde{b} > \bar{b}$ and $f_{\tilde{b}}(0) = f_{\tilde{b}}(x_2^{\tilde{b}})$. Let $b^* = \inf\{b \mid \nabla g(0) > \tilde{b} > \bar{b}, f_b(0) = f_b(x_2^b)\}$. We have*

$$\mathbf{Prox}_g(b) = \operatorname*{argmin}_{x \geq 0} f_b(x) \begin{cases} = x^b, & \text{if } b > \nabla g(0), \\ = x_2^b, & \text{if } \nabla g(0) \geq b > b^*, \\ \ni 0, & \text{if } b \leq b^*. \end{cases} \tag{4}$$

**Remark:** If $b^*$ exists, when $b \leq b^*$, it is possible that $C_b(x)$ and $D(x)$ have more than two intersection points. If $b^*$ does not exist, when $b \leq \nabla g(0)$, it is also possible that $C_b(x)$ and $D(x)$ have more than two intersection points.

*Proof.* Case 1 : Suppose we have $C_{g'(0)}(x) = \nabla g(0) - x \leq \nabla g(x)$ for all $x$ on $(0, \nabla g(0))$. Notice for all $b \leq \nabla g(0)$, we have $\nabla g(x) = D(x) - C_b(x) \geq 0$, so the minimum point of $f_b(x)$ is 0. For all $b > \nabla g(0)$, $C_b = b - x$ and $D(x)$ have only one intersection point denoted as $(x^b, y^b)$. Then, we can easily see that $f_b$ is decreasing on $(0, x^b)$ and increasing on $(x^b, b)$. So, when $b > \nabla g(0)$, the minimum point of $f_b(x)$ is $x^b$.

**Case 2 :** Suppose there exists $0 < \hat{x} < \nabla g(0)$ such that $C_{\nabla g(0)}(\hat{x}) = \nabla g(0) - \hat{x} > \nabla g(\hat{x})$. Then, $D(x)$ and $C_b(x)$ have two intersection points, i.e., $(0, \nabla g(0))$ and $(x_2^{\nabla g(0)}, y_2^{\nabla g(0)})$. It is easily checked that $f_{\nabla g(0)}$ is strictly decreasing on $(0, x_2^{\nabla g(0)})$, so we have $f_{\nabla g(0)}(x_2^{\nabla g(0)}) < f_{\nabla g(0)}(0)$. Also, since $f_{\bar{b}}$ is strictly increasing on $(0, x_2^{\bar{b}})$, we have $f_{\bar{b}}(x_2^{\bar{b}}) > f_{\bar{b}}(0)$.

Because we have
$$\begin{cases} f_{\bar{b}}(x_2^{\bar{b}}) > f_{\bar{b}}(0), \\ f_{\nabla g(0)}(x_2^{\nabla g(0)}) < f_{\nabla g(0)}(0), \end{cases}$$
by a direct computation, we get
$$\begin{cases} g(x_2^{\bar{b}}) - x_2^{\bar{b}} \nabla g(x_2^{\bar{b}}) - \frac{1}{2}(x_2^{\bar{b}})^2 > 0, \\ g(x_2^{\nabla g(0)}) - x_2^{\nabla g(0)} \nabla g(x_2^{\nabla g(0)}) - \frac{1}{2}(x_2^{\nabla g(0)})^2 < 0. \end{cases}$$

So, according to the intermediate value theorem, there exists $\tilde{x}$ such that $g(\tilde{x}) - \tilde{x}\nabla g(\tilde{x}) - \frac{1}{2}(\tilde{x})^2 = 0$. Let $\tilde{b} = \nabla g(\tilde{x}) + \tilde{x}$. Then, $(\tilde{x}, \tilde{b} - \tilde{x})$ is the intersection point of $C_{\tilde{b}}(x)$ and $D(x)$ such that $f_{\tilde{b}}(\tilde{x}) = f_{\tilde{b}}(0)$. Since $x_2^{\bar{b}} < \tilde{x} < x_2^{\nabla g(0)}$ and $\nabla g$ is convex and nonincreasing, we conclude that $\bar{b} < \tilde{b} < \nabla g(0)$. Next, we set $b^* = \inf\{b \mid \bar{b} < \tilde{b} < \nabla g(0), f_b(0) = f_b(x_2^b)\}$.

Given $\nabla g(0) \geq b > \bar{b}$, we can easily see that $f_b$ is increasing on $(0, x_1^b)$, decreasing on $(x_1^b, x_2^b)$ and increasing on $(x_2^b, b)$. So, 0 and $x_2^b$ are two local minimum points of $f_b(x)$ on $[0, b]$.

Next, for $\nabla g(0) \geq b > b^*$, set $b = b^* + \varepsilon$ for some $\varepsilon > 0$. We have
$$\begin{aligned} &f_b(x_2^{b^*}) - f_b(0) \\ =& \frac{1}{2}(x_2^{b^*} - b^* - \varepsilon)^2 + g(x^*) - \frac{1}{2}(b^* + \varepsilon)^2 \\ =& \frac{1}{2}(x_2^{b^*} - b^*)^2 - \frac{1}{2}(b^*)^2 - \varepsilon(x_2^{b^*} - b^*) - \varepsilon b^* \\ =& f_{b^*}(x_2^{b^*}) - f_{b^*}(0) - \varepsilon x_2^* \\ =& -\varepsilon x_2^* < 0. \end{aligned}$$

Since $f_b$ is decreasing on $(x_2^{b^*}, x_2^b)$, we conclude that $f_b(0) > f_b(x_2^{b^*}) \geq f_b(x_2^b)$. So, when $b > b^*$, $x_2^b$ is the global minimum of $f_b(x)$ on $[0, b]$.

Then, for all $\bar{b} < b \leq b^*$, we show that $f_b(0) \leq f_b(x_2^b)$. We prove by contradiction. Suppose that there exists $b$ such that $f_b(0) > f_b(x_2^b)$. Because we have
$$\begin{cases} f_{\bar{b}}(x_2^{\bar{b}}) > f_{\bar{b}}(0), \\ f_b(x_2^b) < f_b(0), \end{cases}$$
by a direct computation, we get
$$\begin{cases} g(x_2^{\bar{b}}) - x_2^{\bar{b}} \nabla g(x_2^{\bar{b}}) - \frac{1}{2}(x_2^{\bar{b}})^2 > 0, \\ g(x_2^b) - x_2^b \nabla g(x_2^b) - \frac{1}{2}(x_2^b)^2 < 0. \end{cases}$$

So, according to the intermediate value theorem, there exists $\tilde{x}_1$ such that $g(\tilde{x}_1) - \tilde{x}_1 \nabla g(\tilde{x}_1) - \frac{1}{2}(\tilde{x}_1)^2 = 0$ and $x_2^{\bar{b}} < \tilde{x}_1 < x_2^b$. Let $\tilde{b}_1 = \nabla g(\tilde{x}_1) + \tilde{x}_1$. Then, $(\tilde{x}_1, \tilde{b}_1 - \tilde{x}_1)$ is the intersection point of $C_{\tilde{b}_1}(x)$ and $D(x)$ such that $f_{\tilde{b}_1}(\tilde{x}_1) = f_{\tilde{b}_1}(0)$. Since $x_2^{\bar{b}} < \tilde{x} < x_2^b$ and $\nabla g$ is convex and nonincreasing, we conclude that $\bar{b} < \tilde{b} < b \leq b^*$, which contradicts the minimality of $b^*$.

Next, when $b \leq \bar{b}$, we have $\nabla f_b(x) = D(x) - C_b(x) \geq 0$, so the global minimum of $f_b(x)$ on $[0, b]$ is 0. Also, when $b > \nabla g(0)$, $C_b = b - x$ and $D(x)$ have only one intersection point $(x^b, y^b)$. Then, we can easily see that $f_b$ is decreasing on $(0, x^b)$ and increasing on $(x^b, b)$. So, when $b > \nabla g(0)$, the global minimum point of $f_b(x)$ is $x^b$. □

## 1.3 Proof of Corollary 1

**Corollary 1.** *Given $g$ satisfying **Assumption 1** in problem (1). Denote $\hat{x}^b = \max\{x | \nabla f_b(x) = 0, 0 \leq x \leq b\}$ and $x^* = \arg\min_{x \in \{0, \hat{x}^b\}} f_b(x)$. Then $x^*$ is optimal to (1), i.e., $x^* \in \mathbf{Prox}_g(b)$.*

*Proof.* As shown in Proposition 2 and 3, when $b$ is larger than a certain threshold, $\mathbf{Prox}_g(b)$ ($x_2^b$ in (2)(4) or $x^b$ in (3)(4)) is unique. Actually the unique solution is the largest intersection point of $C_b(x)$ and $\nabla g(x)$, i.e., $\mathbf{Prox}_g(b) = \hat{x}^b = \max\{x|\nabla f_b(x) = 0, 0 \leq x \leq b\}$. For all the other choices of $b$, $0 \in \mathbf{Prox}_g(b)$. Thus, 0 and $\hat{x}^b$, one of them should be optimal to (1). Thus $x^* = \arg\min_{x \in \{0, \hat{x}^b\}} f_b(x)$ is optimal to (1). □

## 2 Proof of Theorem 2

**Theorem 2.** *For any lower bounded function $g$, its proximal operator $\mathbf{Prox}_g(\cdot)$ is monotone, i.e., for any $p_i^* \in \mathbf{Prox}_g(x_i)$, $i = 1, 2$, $p_1^* \geq p_2^*$, when $x_1 > x_2$.*

*Proof.* The lower bound assumption of $g$ guarantees a finite solution to problem (1). By the optimality of $p_i^*$, $i = 1, 2$, we have

$$g(p_2^*) + \frac{1}{2}(p_2^* - x_1)^2 \geq g(p_1^*) + \frac{1}{2}(p_1^* - x_1)^2, \tag{5}$$

$$g(p_1^*) + \frac{1}{2}(p_1^* - x_2)^2 \geq g(p_2^*) + \frac{1}{2}(p_2^* - x_2)^2. \tag{6}$$

Summing them together gives

$$(p_2^* - x_1)^2 + (p_1^* - x_2)^2 \geq (p_1^* - x_1)^2 + (p_2^* - x_2)^2. \tag{7}$$

It reduces to

$$(p_1^* - p_2^*)(x_1 - x_2) \geq 0. \tag{8}$$

Thus $p_1^* \geq p_2^*$ when $x_1 > x_2$. □

## 3 Convergence Analysis of Algorithm 1

Assume there exists

$$\hat{x}^b = \max\{x|\nabla f_b(x) = \nabla g(x) + x - b = 0, 0 \leq x \leq b\};$$

otherwise, 0 is a solution to (1).

We only need to prove that the fixed point iteration guarantees to find $\hat{x}^b$.

First, if $\nabla g(b) = 0$, then we have found $\hat{x}^b = b$.

For the case $\hat{x}^b < b$, we prove that, the fixed point iteration, starting from $x_0 = b$, converges to $\hat{x}^b$. Indeed, we have

$$b - \nabla g(x) < x, \text{ for any } x > \hat{x}^b.$$

We prove this by contradiction. Assume there exists $\tilde{x} > \hat{x}^b$ such that $b - \nabla g(\tilde{x}) > \tilde{x}$. Notice $g$ satisfies Assumption 1. It is easy to see $\nabla g$ is continuous, decreasing and nonnegative. Then we have $b - \nabla g(b) < b$ ($\nabla g(b) > 0$ since $b > \hat{x}^b$). Thus there must exist some $\hat{x} \in (\min(b, \tilde{x}), \max(b, \tilde{x})) > \hat{x}^b$ such that $b - g(\hat{x}) = \hat{x}$. This contradicts the definition of $\hat{x}^b$.

So, we have

$$x_{k+1} = b - \nabla g(x_k) < x_k, \text{ if } x_k > \hat{x}^b.$$

On the other hand, $\{x_k\}$ is lower bounded by $\hat{x}^b$. So there must exist a limit of $\{x_k\}$, denoted as $\bar{x}$, which is no less than $\hat{x}^b$. Let $k \to +\infty$ on both sides of

$$x_{k+1} = b - \nabla g(x_k),$$

and we see that $\bar{x} = b - \nabla g(\bar{x})$. So, $\bar{x} = \hat{x}^b$, i.e., $\lim_{k \to +\infty} x_k = \hat{x}^b$.

## 4 Convergence Analysis of Generalized Proximal Gradient Algorithm

Consider the following problem

$$\min_{\mathbf{X}} F(\mathbf{X}) = \sum_{i=1}^{m} g(\sigma_i(\mathbf{X})) + h(\mathbf{X}), \tag{9}$$

where $g : \mathbb{R}^+ \to \mathbb{R}^+$ is continuous, concave and nonincreasing on $[0, +\infty)$, and $h : \mathbb{R}^{m \times n} \to \mathbb{R}^+$ has Lipschitz continuous gradient with Lipschitz constant $L(h)$. The Generalized Proximal Gradient (GPG) algorithm solves the above problem by the following updating rule

$$\begin{aligned}\mathbf{X}^{k+1} &= \arg\min_{\mathbf{X}} \sum_{i=1}^m g(\sigma_i(\mathbf{X})) + h(\mathbf{X}^k) + \langle \nabla h(\mathbf{X}^k), \mathbf{X} - \mathbf{X}^k \rangle + \frac{\mu}{2} \|\mathbf{X} - \mathbf{X}^k\|_F^2 \\ &= \arg\min_{\mathbf{X}} \sum_{i=1}^m g(\sigma_i(\mathbf{X})) + \frac{\mu}{2} \|\mathbf{X} - \mathbf{X}^k + \frac{1}{\mu} \nabla h(\mathbf{X}^k)\|_F^2.\end{aligned} \quad (10)$$

Then we have the following results.

**Theorem 3.** *If $\mu > L(h)$, the sequence $\{\mathbf{X}^k\}$ generated by (10) satisfies the following properties:*

*(1) $F(\mathbf{X}^k)$ is monotonically decreasing. Indeed,*

$$F(\mathbf{X}^k) - F(\mathbf{X}^{k+1}) \geq \frac{\mu - L(h)}{2} \|\mathbf{X}^k - \mathbf{X}^{k+1}\|_F^2 \geq 0;$$

*(2) $\lim_{k \to +\infty} (\mathbf{X}^k - \mathbf{X}^{k+1}) = \mathbf{0}$;*

*(3) If $F(\mathbf{X}) \to +\infty$ when $\|\mathbf{X}\|_F \to +\infty$, then any limit point of $\{\mathbf{X}^k\}$ is a stationary point.*

*Proof.* Since $\mathbf{X}^{k+1}$ is optimal to (10), we have

$$\begin{aligned}&\sum_{i=1}^m g(\sigma_i(\mathbf{X}^{k+1})) + h(\mathbf{X}^k) + \langle \nabla h(\mathbf{X}^k), \mathbf{X}^{k+1} - \mathbf{X}^k \rangle + \frac{\mu}{2} \|\mathbf{X}^{k+1} - \mathbf{X}^k\|_F^2 \\ &\leq \sum_{i=1}^m g(\sigma_i(\mathbf{X}^k)) + h(\mathbf{X}^k) + \langle \nabla h(\mathbf{X}^k), \mathbf{X}^k - \mathbf{X}^k \rangle + \frac{\mu}{2} \|\mathbf{X}^k - \mathbf{X}^k\|_F^2 \\ &= \sum_{i=1}^m g(\sigma_i(\mathbf{X}^k)).\end{aligned} \quad (11)$$

On the other hand, since $h$ has Lipschitz continuous gradient, we have [1]

$$h(\mathbf{X}^{k+1}) \leq h(\mathbf{X}^k) + \langle \nabla h(\mathbf{X}^k), \mathbf{X}^{k+1} - \mathbf{X}^k \rangle + \frac{L(h)}{2} \|\mathbf{X}^{k+1} - \mathbf{X}^k\|_F^2. \quad (12)$$

Combining (11) and (12) leads to

$$\begin{aligned}&F(\mathbf{X}^k) - F(\mathbf{X}^{k+1}) \\ &= \sum_{i=1}^m g(\sigma_i(\mathbf{X}^k)) + h(\mathbf{X}^k) - \sum_{i=1}^m g(\sigma_i(\mathbf{X}^{k+1})) - h(\mathbf{X}^{k+1}) \\ &\geq \frac{\mu - L(h)}{2} \|\mathbf{X}^{k+1} - \mathbf{X}^k\|_F^2.\end{aligned} \quad (13)$$

Thus $\mu > L(h)$ guarantees that $F(\mathbf{X}^k) \geq F(\mathbf{X}^{k+1})$.

Summing (13) for $k = 1, 2, \cdots$, we get

$$F(\mathbf{X}^1) \geq \frac{\mu - L(h)}{2} \sum_{k=1}^{+\infty} \|\mathbf{X}^{k+1} - \mathbf{X}^k\|_F^2. \quad (14)$$

This implies that

$$\lim_{k \to +\infty} (\mathbf{X}^k - \mathbf{X}^{k+1}) = \mathbf{0}. \quad (15)$$

Furthermore, since $F(\mathbf{X}) \to +\infty$ when $\|\mathbf{X}\|_F \to +\infty$, $\{\mathbf{X}^k\}$ is bounded. There exist $\mathbf{X}^*$ and a subsequence $\{\mathbf{X}^{k_j}\}$ such that $\lim_{j \to +\infty} \mathbf{X}^{k_j} = \mathbf{X}^*$. By using (15), we get $\lim_{j \to +\infty} \mathbf{X}^{k_j+1} = \mathbf{X}^*$. Considering that $\mathbf{X}^{k_j}$ is optimal to (10), and $-\sum_{i=1}^m g(\sigma_i(\mathbf{X}))$ is convex (since $g$ is concave) [3], there exists $\mathbf{Q}^{k_j+1} \in -\partial \left(-\sum_{i=1}^m g(\sigma_i(\mathbf{X}^{k_j+1}))\right)$ such that

$$\mathbf{Q}^{k_j+1} + \nabla h(\mathbf{X}^{k_j}) + \mu(\mathbf{X}^{k_j+1} - \mathbf{X}^{k_j}) = \mathbf{0}. \quad (16)$$

Let $j \to +\infty$ in (16). By the upper semi-continuous property of the subdifferential [2], there exists $\mathbf{Q}^* \in -\partial\left(-\sum_{i=1}^{m} g(\sigma_i(\mathbf{X}^*))\right)$, such that

$$\mathbf{0} = \mathbf{Q}^* + \nabla h(\mathbf{X}^*) \in \nabla F(\mathbf{X}^*). \tag{17}$$

Thus $\mathbf{X}^*$ is a stationary point to (9). □